\documentclass[runningheads]{llncs}

\usepackage[mobile]{eccv}

\usepackage{eccvabbrv}

\usepackage{graphicx}
\usepackage{booktabs}

\usepackage[accsupp]{axessibility}  

\usepackage[pagebackref,breaklinks,colorlinks,citecolor=eccvblue]{hyperref}

\usepackage{orcidlink}
\newcommand{\method}{Wear-Any-Way\xspace}

\newcommand{\ve}[1]{\mathbf{#1}} 
\usepackage[capitalize]{cleveref}  
\crefname{section}{Sec.}{Secs.}
\Crefname{section}{Section}{Sections}
\crefname{table}{Tab.}{Tabs.}
\Crefname{table}{Table}{Tables}
\crefname{figure}{Fig.}{Figs.}
\Crefname{figure}{Figure}{Figures}
\crefname{equation}{Eq.}{Eqs.}
\Crefname{equation}{Equation}{Equations}

\begin{document}

\title{ Wear-Any-Way: Manipulable Virtual Try-on via Sparse Correspondence Alignment } 

\titlerunning{Wear-Any-Way}


\author{
    Mengting Chen \quad  
    Xi Chen \quad
    Zhonghua Zhai \quad
    Chen Ju \quad
    Xuewen Hong \quad
    Jinsong Lan \quad
    Shuai Xiao\thanks{Corresponding author} \\[5pt]
}
\vspace{-25pt}
\institute{Alibaba Group \\
\email{\{mengtingchen.cs@gmail.com  shuai.xsh@alibaba-inc.com\} }\\
\textcolor{magenta}{ \href{https://mengtingchen.github.io/wear-any-way-page/}{mengtingchen.github.io/wear-any-way-page}}
}

\authorrunning{Chen et al.}


\maketitle

\begin{abstract}
This paper introduces a novel framework for virtual try-on, termed \method.
Different from previous methods, \method is a customizable solution. Besides generating high-fidelity results, our method supports users to precisely manipulate the wearing style. 
To achieve this goal, we first construct a strong pipeline for standard virtual try-on, supporting single/multiple garment try-on and model-to-model settings in complicated scenarios.
To make it manipulable, we propose sparse correspondence alignment which involves point-based control to guide the generation for specific locations. 
With this design, Wear-Any-Way gets state-of-the-art performance for the standard setting and provides a novel interaction form for customizing the wearing style.
For instance, it supports users to drag the sleeve to make it rolled up, drag the coat to make it open, and utilize clicks to control the style of tuck, \textit{etc.} \method enables more liberated and flexible expressions of the attires, holding profound implications in the fashion industry.

\end{abstract}
\section{Introduction}

\begin{figure*}[t]
\centering 
    \includegraphics[width=1.0\linewidth]{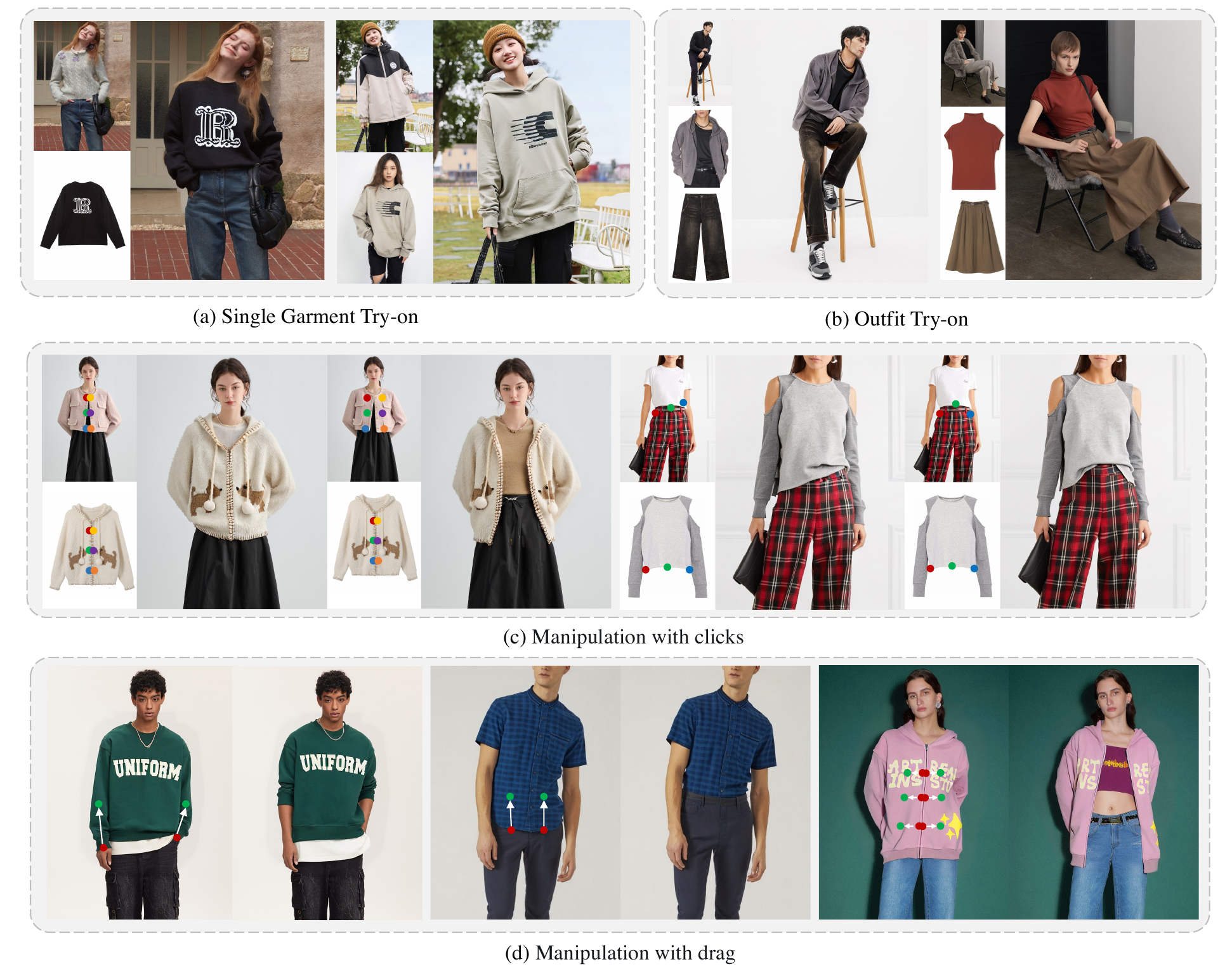}
    \vspace{-18pt}
    \captionsetup{type=figure}
    \caption{%
        \textbf{Manipulable try-on with \method.}
        Our method achieves state-of-the-art performance for the standard setting of virtual try-on~(first row), supporting diversified input formats and scenarios. 
        An impressive feature is that our method supports users to manipulate the way of wearing using simple interactions like click~(second row) and drag~(third row).
        It should be noted that all these applications are accomplished with a single model in one pass.
    }
    \label{fig:teaser}
\vspace{-5pt}
\end{figure*}

Virtual try-on aims to synthesize an image of the specific human wearing the provided garments. It emerges as a significant technology within the realm of the fashion industry, providing consumers with an immersive and interactive means of experiencing apparel without physically trying. 

Classical solutions for virtual try-on rely on Generative-Adversarial-Networks (GANs)~\cite{liu2020mgcm, liu2019toward, dong2020fashion, jo2019sc, lee2020maskgan} to learn the feature warping from the garment image to the person image.  
Recent advances in diffusion models~\cite{ddpm} bring a huge improvement in the generation quality for various image/video synthesis tasks, including virtual try-on. Some recent works~\cite{zhu2023tryondiffusion,kim2023stableviton} could generate high-fidelity results leveraging pre-trained text-to-image diffusion models~\cite{ldm}.
However, there still exist drawbacks for these solutions.

First, most of them only support simple cases like well-arranged single garments with simple textures as input. For garments with complicated textures or patterns, existing solutions often fail to maintain the fidelity of the fine details.  
In addition, most of the previous solutions do not solve the challenges in real-world applications like model-to-model try-on, multi-garment try-on, complex human poses, and complicated scenarios, \textit{etc.}

Second, previous methods are unable to exert control over the wearing style. However, the actual wearing style holds significant importance in the realm of fashion. For instance, variations in the rolling up or down of sleeves, the tucking and layering of tops and bottoms, the decision to leave a jacket open or closed, and even the exploration of different sizes for the same garment all contribute to diverse ways of wearing. These distinct styles can showcase varied states of the same piece of clothing, highlighting the pivotal role of styling options, particularly in the context of fashion applications.

In this work, we propose \method, a novel framework for virtual try-on that solves the two aforementioned challenges at once. 
\method could act as a strong option for the standard setting of virtual try-on, it synthesizes high-quality images and preserves the fine details of the patterns on the garment. 
Besides, it serves as a universal solution for real-world applications, supporting various sub-tasks like model-to-model try-on, multi-garment try-on, and complicated scenarios like street scenes. 
The most important feature is that \method supports users in customizing the wearing style.  As shown in \cref{fig:teaser}, users could use simple interactions like click and drag to control the rolling of sleeves, the open magnitude of the coat, and even the style of tuck. 

To archive these functions, we first build a strong baseline for virtual try-on. Recently, Reference-only~\cite{Reference-only} structure has proven effective in many downstream tasks like image-to-video~\cite{hu2023animateanyone, xu2023magicanimate} and image editing~\cite{cao2023masactrl, mou2023dragondiffusion, nam2024dreammatcher}. Inspired by these methods, we build a dual-branch pipeline, the main branch is a denoising U-Net initialized from a pre-trained inpainting Stable Diffusion~\cite{ldm}. The main U-Net takes the person image as input while the reference U-Net extracts the features from the garment image. The referential garment features are then injected into the main branch with self-attention. To further improve the flexibility and robustness of this strong baseline, we also inject the guidance of human pose. This pipeline achieves state-of-the-art performance in the standard try-on setting.

Afterward, we take a further step to make this strong baseline customizable. Specifically, we investigate a point-based control that forces the specific points on the garment image to match the target points on the person image in the generation result. To align the features of the paired points, we propose sparse correspondence alignment, which first learns a series of permutable point embeddings and injects these embeddings into both the main and reference U-Net by modifying the attention layers. To assist the network learn the feature alignment better, we design several strategies like condition dropping, zero-initialization, and point-weighted loss to ease the optimization.

Equipped with all these techniques, \method demonstrates superior quality and controllability for virtual try-on. 
In general, our contributions could be summarized in three folds:
\begin{itemize}
    \item We construct a novel framework, \method, which generates high-quality results and supports users to precisely manipulate the way of wearing.  
    \item We propose a strong, flexible, and robust baseline for virtual try-on, which reaches state-of-the-art with extensive comparisons with previous methods.
    \item We design the sparse correspondence alignment to enable the point-based control and further develop several strategies~(\textit{i.e.}, conditional dropping, zero-initialization, point-weighted loss) to enhance the controllability.
\end{itemize}
\vspace{-2mm}
\section{Related Work}
\vspace{-2mm}
\noindent \textbf{GAN-based virtual try-on.} 
To execute the Virtual try-on task, numerous methods~\cite{liu2020mgcm, liu2019toward, dong2020fashion, jo2019sc, lee2020maskgan} have utilized Generative Adversarial Networks (GANs). Effectively applying GANs to the virtual try-on challenge requires a nuanced approach, current methods \cite{choi2021viton,lee2022high,xie2023gp,ge2021parser} typically adopt a two-stage strategy: first deforming the garment to fit the target body shape, then integrating the transformed clothing onto the human model using a GAN-based try-on generator. To achieve accurate deformation, several techniques estimate a dense flow map that guides the clothing reshaping process \cite{lee2022high,ge2021parser,han2019clothflow,bai2022single,xie2023gp}. Subsequent methods then address misalignment issues between the warped clothing and the human body using strategies like normalization \cite{choi2021viton} or distillation \cite{issenhuth2020not, ge2021parser}. However, these existing approaches have limitations, particularly when dealing with images of individuals in complex poses or against intricate backgrounds, resulting in noticeable drops in performance. Moreover, Conditional GANs (cGANs) encounter difficulties with significant spatial transformations between the clothing and the subject's posture, as highlighted by CP-VTON \cite{wang2018toward}, which brings to light the need for improved methods capable of handling these challenges.

\noindent \textbf{Diffusion-based virtual try-on.} 
The exceptional generative capabilities of diffusion have inspired several approaches to incorporate diffusion models into fashion synthesis, covering tasks like visual try-on \cite{bhunia2023person, cao2023difffashion, karras2023dreampose, morelli2023ladi, zhu2023tryondiffusion}. TryOnDiffusion \cite{zhu2023tryondiffusion} utilizes dual U-Nets for the try-on task. However, the requirement for extensive datasets capturing various poses presents a significant challenge. Consequently, there has been a pivot towards leveraging large-scale pre-trained diffusion models as priors in the try-on process \cite{ho2020denoising,rombach2022high,radford2021learning,yang2023paint}. Approaches like LADI-VTON \cite{morelli2023ladi} and DCI-VTON \cite{gou2023taming} have been introduced, which treat clothing as pseudo-words or use warping networks to integrate garments into pre-trained diffusion models. StableVITON \cite{kim2023stableviton} proposes a novel approach that conditions the intermediate feature maps of a spatial encoder using a zero cross-attention block. While these methods have addressed the issues of background complexity, they struggle to preserve fine details and encounter problems such as inaccurate warping. Moreover, they fail to enable flexible control over how garments are worn, producing only a rigid, static image.

Building on this landscape, our work extend the interactive capabilities of virtual try-on, which sets new standards for performance and user interaction. Rather than just generating static images, \method empowers users to manipulate garments dynamically, allowing for an unprecedented level of customization that signifies a significant leap in the personalization of digital fashion experiences.

\noindent \textbf{Point-based image editing.} 
Building on the achievements of diffusion models~\cite{ho2020denoising}, various diffusion-based image editing techniques~\cite{avrahami2022blended, p2p, kawar2023imagic, meng2021sdedit, brooks2023instructpix2pix} have emerged, predominantly relying on textual instructions for editing. Techniques such as those in \cite{kawar2023imagic, valevski2023unitune, kwon2022diffusion} apply fine-tuning to models on single images to produce alterations directed by descriptive texts. However, this text-guided approach often yields only broad-stroke modifications, lacking the precision required for detailed image editing.

To overcome the limitations of text-guided editing, studies explored point-based editing \cite{pan2023drag, endo2022user, wang2022rewriting}. DragGAN, notable for its intuitive drag-and-drop manipulation, optimizes latent codes for handle points and incorporates point tracking. However, GANs' inherent limitations constrain DragGAN. FreeDrag \cite{ling2023freedrag} refines DragGAN by eliminating point tracking, while \cite{shi2023dragdiffusion} extends DragGAN's framework to diffusion models, showcasing versatility. Simultaneously, \cite{mou2023dragondiffusion} utilizes diffusion models for drag-based editing, employing classifier guidance to convert editing intentions into actionable gradients.

To address these shortcomings and enhance the granularity and adaptability of image editing, our research leverages diffusion models' exceptional generative capabilities. We propose a novel editing paradigm, \method, combining point-based precision with diffusion models' rich generative potential. \method enhances fine-grained, context-aware image alterations, offering unprecedented control over the editing process and outcomes.
\section{Method}
We first introduce the basic knowledge required for diffusion models in \cref{sec:preliminaries}. Afterward, we present our strong baseline for virtual try-on in \cref{sec:baseline}. Next, we dive into the details of our proposed sparse correspondence alignment in \cref{sec:link} and the training strategies in \cref{sec:strategies}. 
In addition, in \cref{sec:collect} we also elaborate on the details for collecting the training point pairs and finally summarize the the inference pipeline in \cref{sec:ui}.

\subsection{Preliminaries} \label{sec:preliminaries}
\noindent\textbf{Text-to-image diffusion model.} Diffusion models~\cite{ddpm} exhibit promising capabilities in both image and video generation. In this study, we select the widely adopted Stable Diffusion~\cite{ldm} as our foundational model, leveraging its efficient denoising procedure in the latent space.
The model initially employs a latent encoder~\cite{VAE} to project  an input image $\ve{x}_0$ into the latent space: $\ve{z}_0 = \mathcal{E}(\ve{x}_0)$. Throughout training, Stable Diffusion transforms the latent representation into Gaussian noise using the formula:

\begin{equation}
\label{eq:add_noise}
\ve{z}_t = \sqrt{\bar{\alpha_t}} \ve{z}_0 + \sqrt{1 - \bar{\alpha_t}} \ve{\epsilon},
\end{equation}

where $\ve{\epsilon} \sim \mathcal{U}([0, 1])$, and $\bar{\alpha_t}$ is the cumulative product of the noise coefficient $\alpha_t$ at each step. Subsequently, the model learns to predict the added noise as:

\begin{equation}
\label{euq:ldm}
\mathbb{E}_{\ve{z}, \ve{c}, \ve{\epsilon}, t}(| \ve{\epsilon}_{\theta}(\ve{z}_t, \ve{c}, t) - \ve{\epsilon} |^2_2).
\end{equation}

Here, $t$ represents the diffusion timestep, and $\ve{c}$ denotes the conditioning text prompts. During the inference phase, Stable Diffusion effectively reconstructs an image from Gaussian noise step by step, predicting the noise added at each stage. The denoised results are then fed into a latent decoder to regenerate colored images from the latent representations, denoted as $\ve{\hat x}_0 = \mathcal{D}(\ve{\hat z}_0)$.

\begin{figure}[t]
\centering 
\includegraphics[width=0.98\linewidth]{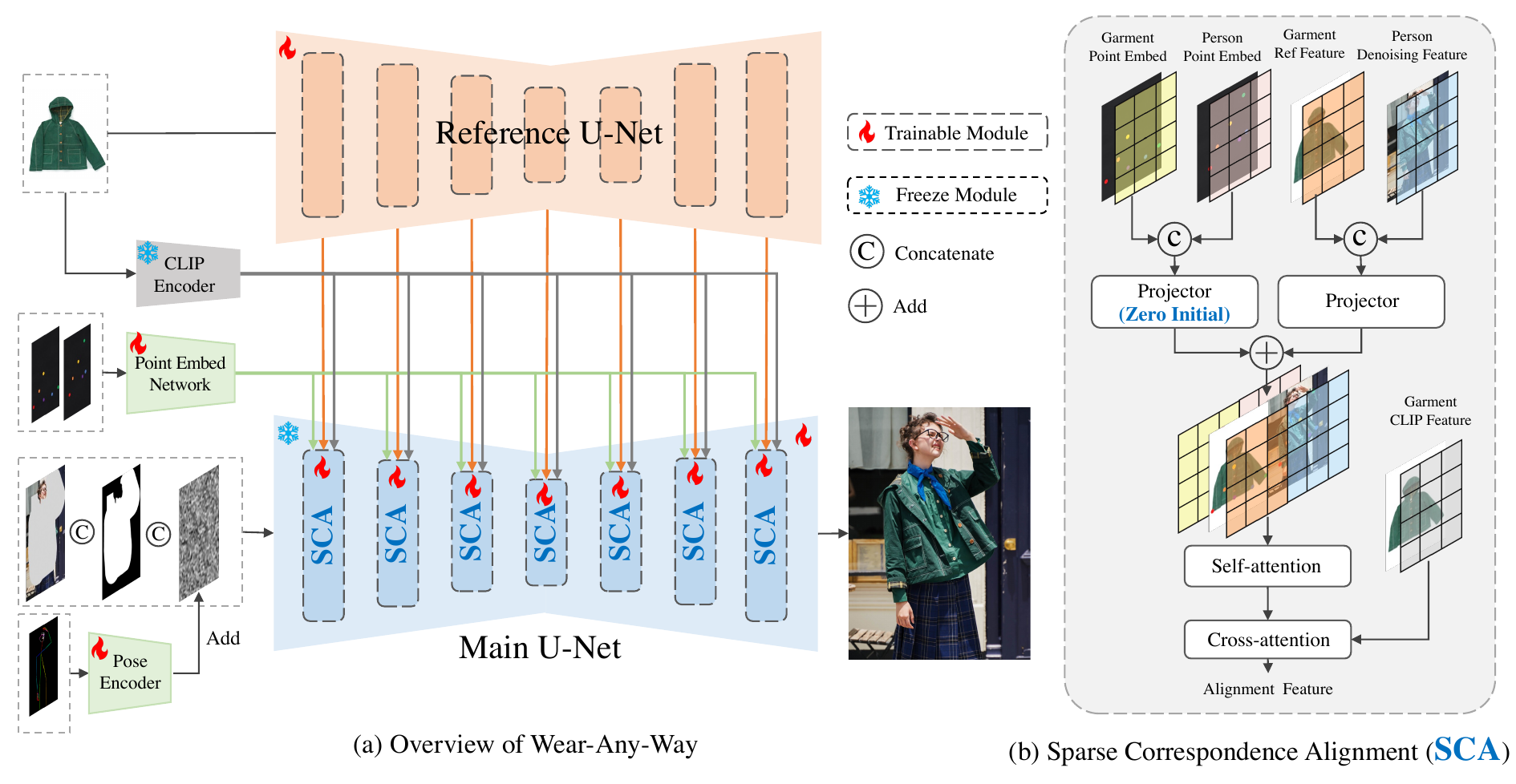} 
\vspace{-5pt}
\caption{%
    \textbf{The pipeline of Wear-Any-Way.} 
    The overall framework consists of two U-Nets. 
    The reference U-Net takes the garment image as input to extract fine-grained features.
    The main U-Net takes charge of generating the try-on results. It takes the person image~(masked), the garment mask, and the latent noise as input. We exert the pose control via an additional pose encoder. The point-based control is realized by a point embedding network and a sparse correspondence alignment module. The detailed structures are demonstrated on the right part. 
    Symbols of flames and snowflakes denote trainable and frozen parameters respectively.
}
\label{fig:main_pipeline}
\vspace{-15pt}
\end{figure}

\subsection{Virtual Try-on Pipeline} \label{sec:baseline}
As demonstrated in \cref{fig:main_pipeline}, our pipeline consists of two branches. The main branch is the inpainting model initialized with the pre-trained weight of Stable Diffusion~\cite{ldm}. It takes in a 9-channel tensor as input, with 4 channels of latent noise, 4 channels of the latent for the inpainting background~(\textit{i,e.,} person image with the masked clothes region), and 1 channel for the binary mask~(representing for inpainting region). The original SD receives text embedding as conditions to guide the diffusion procedure. Instead, we replace the text embedding with an image embedding of the garment image extracted by a CLIP~\cite{CLIP} image encoder. 

The CLIP image embedding could guarantee the overall colors and textures of the garment but fails to preserve the fine details. 
Recently, reference-only~\cite{Reference-only} has proven effective in keeping the fine details of the reference image in many fields of applications~\cite{xu2023magicanimate, hu2023animateanyone,cao2023masactrl,nam2024dreammatcher}. TryonDiffusion~\cite{zhu2023tryondiffusion} also leverages two U-Net to make generations with high fidelity. Inspired by those explorations, we also use a reference U-Net to extract the detailed features of the garment. Our reference  U-Net is a standard text-to-image diffusion model with 4-channel input. We conduct feature fusion after each block by concatenating the ``key'' and ``value'' of the reference U-Net after the main U-Net. 

To further enhance the generation, we add the pose map as an additional control. We construct a tiny convolution network to extract the features of the pose map, and directly add it to the latent noise of the main U-Net. The pose map is extracted from the provided person image with DW-Pose~\cite{DWPose}.

\subsection{Sparse Correspondence Alignment} \label{sec:link}
To make our \method customizable, we introduce a sparse correspondence alignment mechanism into the diffusion procedure. Specifically, for the point-pair, one point is marked on the garment image and another is given on the person image. We utilize the correspondence between the two points to control the generation results: the marked position of the garment would match the targeted position on the person image. In this way, users could precisely control the wearing style by manipulating multiple point pairs. 

As shown in \cref{fig:main_pipeline}, we first learn a series of point embedding to represent the pair of control points. Afterward, we inject this control signal into both the main U-Net and referential U-Net. To assist the model in learning the correspondence relationships, we propose several strategies like condition dropping, zero-initialization, and point-guided loss.  

\vspace{5pt}
\noindent\textbf{Point embedding.}  Assuming we sample N pairs of points, we use disk maps $\mathbf{D}_{g/p}^{1 \times H \times W}$  to represent the points on the garment and person images respectively. 
The background values are zeros. Points on the garment image are filled with values from 1 to K randomly without repeat, while the points on the person image are filled with the corresponding values. 
K denotes the maximum number of control points enabled in one image with $N \leq K$.  In this work, we set K = 24.
This random assignment decouples the semantics and the points thus making the point representation permutable. It serves as the basis to support arbitrary numbers of point control at arbitrary locations.

Afterward, we design a point embedding network with stacked convolution layers to project the disk maps into high-dimension embeddings:
$\mathbf{E}_{g/p}^{C \times H \times W}$. This network is optimized alone with the diffusion model in the end-to-end training. 

\vspace{5pt}
\noindent\textbf{Embedding injection.} 
We excel in controls by injecting the point embedding into the attention layer. In our baseline, the features from the reference U-Net are concatenated on the ``key'' and ``value'' of the self-attention as illustrated in \cref{eq:reference}, where the subscripts of m,r represents the main and reference U-Net.

\begin{equation}
   \mathrm{Attention} = \mathrm{softmax}(\frac{Q_{m} \cdot \mathrm{cat}(K_{m}, K_{r}) ^T}{\sqrt{d_k}})  \cdot \mathrm{cat}(V_m,V_r)
   \label{eq:reference}
\end{equation}

This attention layer enables garment features extracted by the reference U-Net to be integrated into the main U-Net. 
To enable the correspondence control with point guidance, we modify this attention layer via adding the point embedding of the person and garment with the ``query'' and ``key'' as in \cref{eq:point}.

\begin{equation}
   \mathrm{Attention} = \mathrm{softmax}(\frac{ ({Q_{m}\!+\! \color{blue}{E_{p}}}) \cdot \mathrm{cat}({K_{m}\!+\!{\color{blue}E_{p}}, K_{r}\!+\!{\color{blue}E_{g}}) ^T}}{\sqrt{d_k}})  \cdot \mathrm{cat}(V_m,V_r)
   \label{eq:point}
\end{equation}

In this way, when integrating the garment feature into the main U-Net, the feature aggregation would consider the correspondence of the point pairs. 
The feature located by the point of the garment could be aligned to the position of the point on the person image. Thus, 
users could assign control points to manipulate the wearing style by clicking and dragging.

\subsection{Training strategies} \label{sec:strategies}
\vspace{5pt}
Besides the design of the model structure, we also develop several training strategies to assist \method to learn the correspondence alignment.

\noindent\textbf{Condition dropping.} 
We observe that, even without the guidance of point pairs, the generation results on the training set are already very close to the ground truth.
We analyze that, the inpainting mask and the human pose map could indicate the wearing style of the training sample to some extent. 
To enforce the model to learn from the control point, we increase the possibility of dropping the pose map and degrading the inpainting mask to a box around the ground truth mask.  

\noindent\textbf{Zero-initialization.} Adding the point embedding to the attention ``key'' and ``value'' causes the unstability of the training optimization.  To achieve a progressive integration, we get the inspiration of ControlNet~\cite{contolnet} to add a zero-initialized convolutional layer at the output of the point embedding network. This zero-initialization brings better convergence.

\noindent\textbf{Point-weighted loss.} To enhance the controllability of the paired points. We increase the loss weight around the sampled points on the person image. The supervision of \method is an MSE loss for predicting the noise.

\begin{figure}[t]
\centering 
\includegraphics[width=0.9\linewidth]{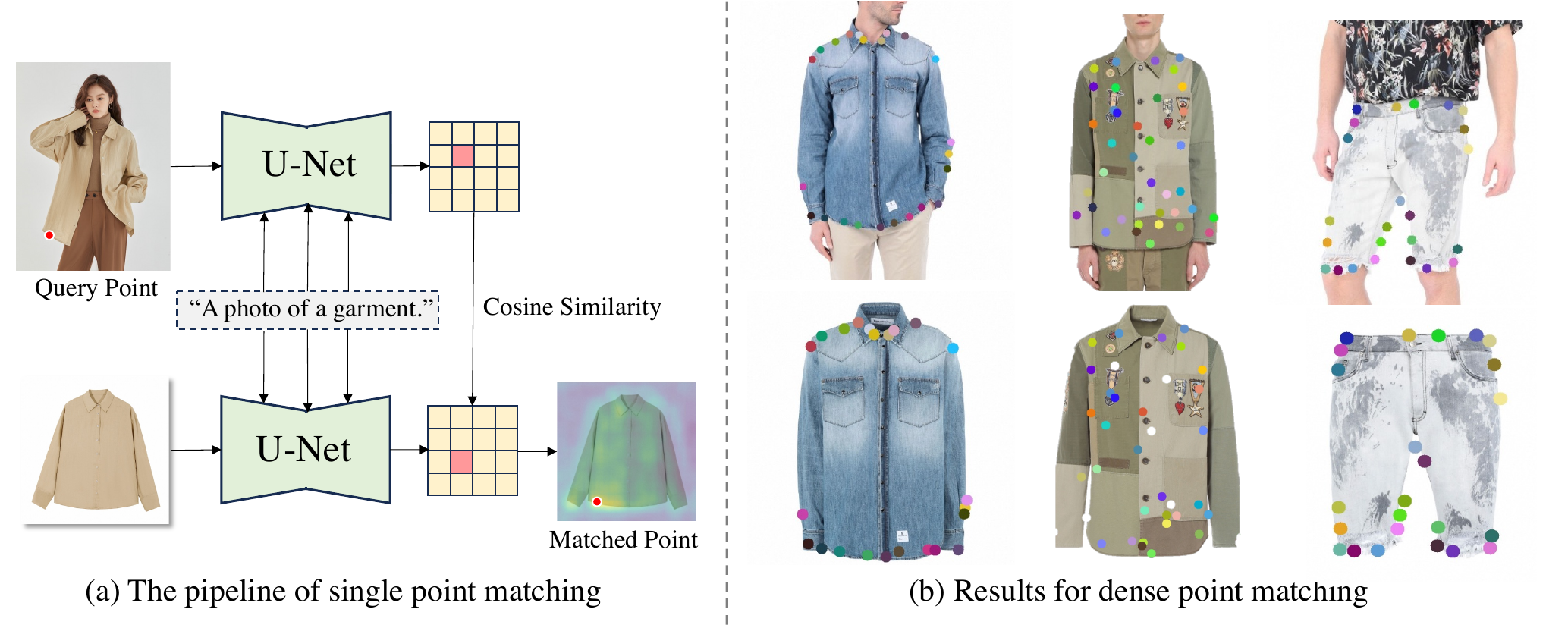} 
\vspace{-5pt}
\caption{%
    \textbf{Pipeline of collecting the training point-pairs.} 
    As shown on the left, the person and garment images are sent into the same Stable Diffusion to extract the feature. We calculate the cosine similarity between the two feature maps to get the point pairs. Some densely sampled point pairs are demonstrated on the right.
}
\label{fig:point_collection}
\vspace{-15pt}
\end{figure}

\subsection{Training Points Collection} \label{sec:collect}
In this section, we elaborate on the pipeline for collecting the training point pairs. 
It is crucial to get precisely matched point pairs from the garment and person image to train \method. Considering that there are no densely annotated point data between the garment and person image, we make extensive explorations to collect point pairs. 

The challenge lies in the fact that garments are not rigid like steel. When worn on the human body, garments could undergo deformation. Previous virtual matching/correspondence learning methods~\cite{sarlin2020superglue,pautrat2023gluestick,lindenberger2023lightglue} could only deal with rigid objects like buildings. 
Fashion/human key points detection methods~\cite{zou2019fashionai,liu2021mmfashion, DWPose} could localize point pairs. However, they could only detect a few predefined key points, which fail to generate arbitrary sampled points for truly flexible control.
Recently, some works~\cite{hedlin2024unsupervisedmatcher,zhang2024tale,tang2024emergent} find that the pre-trained diffusion models are naturally image matchers. In this work, we also explore correspondence learning by leveraging pre-trained text-to-image diffusion models.

As illustrated in \cref{fig:point_collection}, we leverage the siamese text-to-video diffusion model to extract features from the person and garment image respectively. We take the feature map of the last later and ensemble the prediction results at multiple time steps to get a robust matching. Given a point on the person image, we select the corresponding point on the garment image with the maximum cosine similarity. 
The dense point-matching results are demonstrated on the right of \cref{fig:point_collection}. Given a person image, we first extract the mask for the wearing clothes. Afterward, we randomly sample points in the internal and boundary regions of the mask as queries and leverage the matching pipeline to extract the corresponding points on the garment image. 
We chose the mapping direction from the peron image to the garment image because some points on the garment image could not be matched on the on-body image when the poses were complex.

\begin{figure}[t]
\centering 
\includegraphics[width=0.8\linewidth]{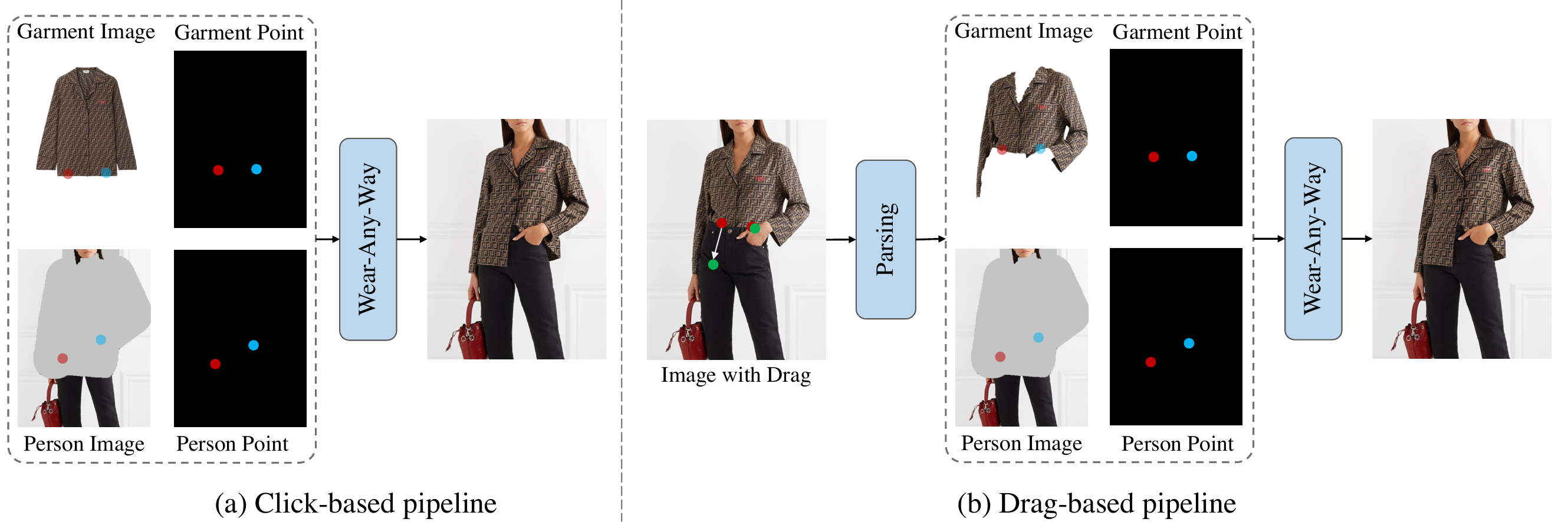} 
\vspace{-5pt}
\caption{%
    \textbf{The inference pipeline of \method. } 
    For click-based control, users provide garment images, person images, and point pairs to customize the generation. 
    When the user drags the image, the starting and end points are translated as the garment and person points. While the parsed clothes are regarded as the garment image. Thus, the drag could be transformed into the click-based setting.
}
\label{fig:point_infer}
\vspace{-5pt}
\end{figure}

\subsection{Inference with manipulation} \label{sec:ui}
Equipped with the sparse correspondence alignment, \method supports users to customize the try-on results using control points. The inference pipeline is illustrated in \cref{fig:point_infer}. For the click-based setting, besides providing the garment image and person images, users could assign multiple point pairs on these two images as control signals. The coordinates on the garment image indicated by the garment points could be aligned to the corresponding position of the person points in the generation result.
For the drag-based control, the starting and end points are processed as garment points and person points, while the parsed clothes are regarded as the garment image. In this way, the drag-based manipulation could be transformed into click-based controls.

\section{Experiments}

\subsection{Implementation Details}
\noindent \textbf{Detailed configurations.} In this work, the main U-Net and the reference U-Net both leverage the pre-trained weights from Stable Diffusion-1.5~\cite{ldm}. We collect 0.3 million high-quality try-on data with ``person image, up-clothes, down-clothes'' triplets to train our model.  However, for fair comparisons with other works, we also train \method on VITON-HD~\cite{vitonhd} and Dresscode~\cite{morelli2022dress} respectively to report the quantitative results and make qualitative comparisons. 
To equip our model with the ability to try the upper and down clothes in one pass. We concatenate two input garment image together from $\text{H} \times \text{W} \times 3 $ to $\text{H} \times 2\text{W} \times 3 $. We randomly drop a garment image to an all-zeros image to preserve the ability of single-garment generation.

\noindent \textbf{Training hyper-parameters.}  During training, we set the initial learning rate 5e-5 with a batch size of 64. The models are trained using $8 \times$ A100 GPUs.  For the main U-Net, we train the parameters of the decoder, and the self-attention layers of the encoder; All parameters of the reference U-Net are trained.
For the data augmentation, we conduct random crop on the person image, and exert random flip simultaneously on the garment and person images. In addition, we also use random color jitter to improve the robustness. We train our model with the resolution of $768\times576$ on the self-collected data for better visual quality. We also train a $512\times384$ version for fair comparisons with previous works.

\subsection{Evaluation protocols.}
\label{sec:eval}
\noindent \textbf{Standard virtual try-on.}
We first evaluate the performance of \method on standard virtual try-on benchmarks~(\textit{i.e.,}VITON-HD~\cite{vitonhd}  Dresscode~\cite{morelli2022dress}) to report qualitative results. Meanwhile, we give qualitative comparisons with state-of-the-art methods to prove the effectiveness of our design.

\noindent \textbf{Evaluation for point control.}
To evaluate the ability of point-based control, we 
get inspiration from previous drag-based image editing methods~\cite{pan2023draggan,shi2023dragdiffusion,mou2023dragondiffusion} to calculate the landmark distance. Specifically, we detect the fashion landmarks using FashionAI~\cite{zou2019fashionai} detector on the pair of garment image and the person image. Afterward, we use the paired landmarks $\mathcal{L}_{garment}$ and $\mathcal{L}_{person}$ as the control points to generate a try-on image~(the person image could be viewed as the ground truth). Next, we use the same detector to localize the landmarks on the newly generated image, noted as $\mathcal{L}_{gen}$. We calculate the Euclidean distance between $\mathcal{L}_{person}$ and $\mathcal{L}_{gen}$ to evaluate the control ability. Ideally, the landmark distance should be small if the generation is well-controlled by the points. We construct a benchmark covering the upper-, down-, and coat-clothes with 1000 samples in total for a comprehensive evaluation.

\begin{figure}[t]
\centering 
\includegraphics[width=0.7\linewidth]{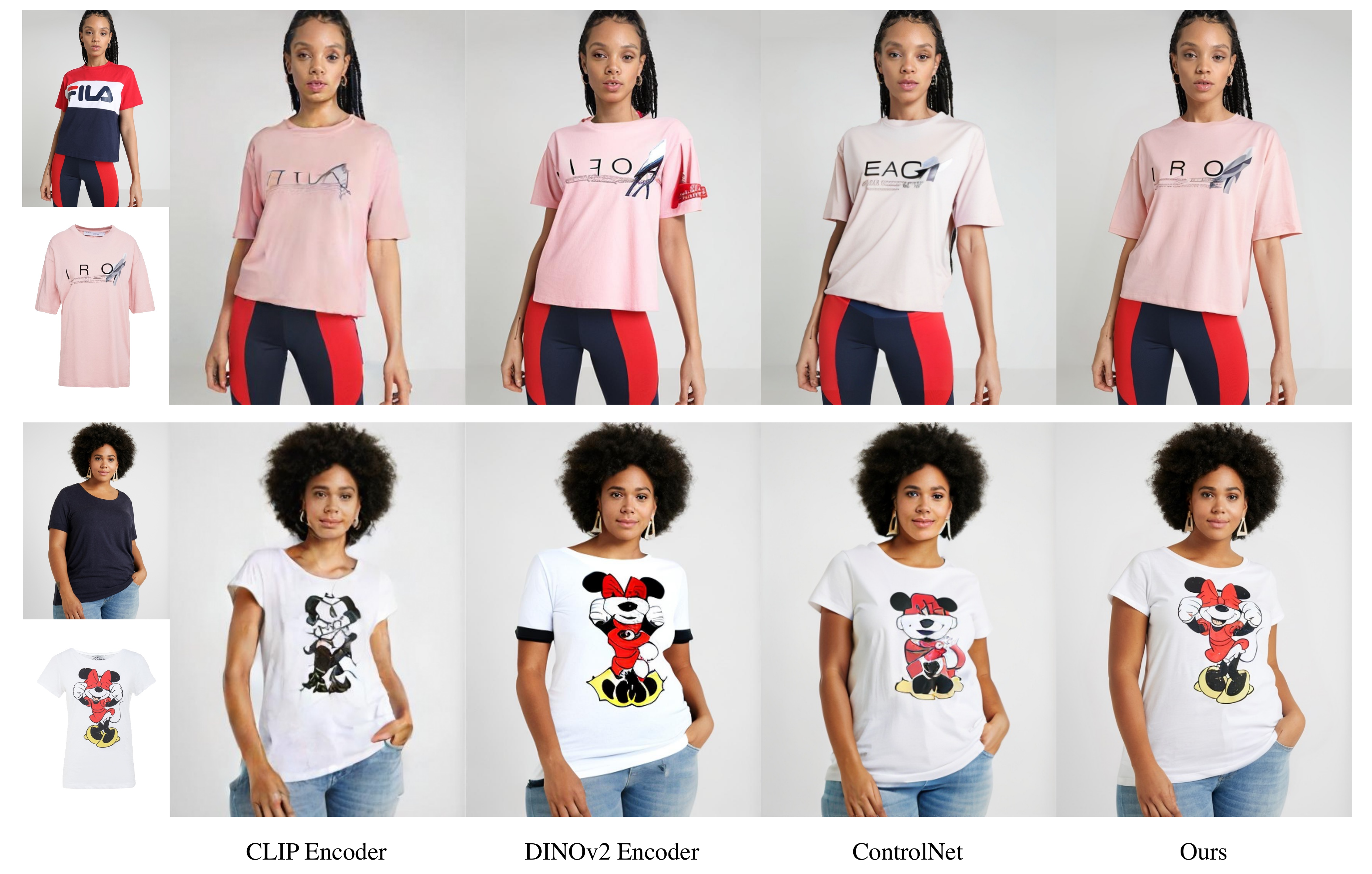} 
\vspace{-5pt}
\caption{%
    \textbf{Ablation studies on different feature extractors.} We compare CLIP image encoder~\cite{CLIP}, DINOv2~\cite{dinov2}, ControlNet~\cite{contolnet}, and our reference U-Net.  Our results demonstrate notable superiority. 
}
\label{fig:tryon_ablation}
\vspace{-5pt}
\end{figure}

\subsection{Ablations Studies}
In this section, we dive into the details to prove the effectiveness of our pipeline design. We first conduct experiments for our strong baseline of standard virtual try-on. Afterward, we provide a detailed analysis of the sparse correspondence alignment module. Besides, collecting the paired points training is also an important step. Thus, we make comparisons with multiple correspondence matching methods with qualitative results. 

\noindent \textbf{Strong baseline.}
We first investigate the design of our strong baseline for virtual try-on. We claim that the reference U-Net is crucial for preserving the fine details of garments.
Previous works like AnyDoor~\cite{chen2023anydoor} uses a image encoder~(\textit{e.g.}, DINOv2~\cite{dinov2}) to extract the garment features. StableVITON~\cite{kim2023stableviton} leverages a ControlNet-like~\cite{contolnet} structure to extract finer representations.

We organize the qualitative comparisons in \cref{fig:tryon_ablation}.  We leverage the CLIP image encoder, DINOv2, and ControlNet as feature extractors and apply the same training settings for fair comparison. We observe that CLIP, DINOv2, and ControlNet could only encode the global appearances of the garments, but fail to preserve the identity of the detailed patterns/texts/logos. 
In contrast, the reference U-Net provides fine-grained details and is able to preserve the high-fidelity details of the garments.

\noindent \textbf{Sparse correspondence alignment.} It is the core component of our point-based control. We first conduct ablation studies for the control injection methods in \cref{tab:embedding_injection}. We follow the evaluation protocol introduced in \cref{sec:eval} to calculate the landmark distance.  Without the point embedding, our try-on baseline~(row~1) gets a high landmark distance. In the second row,  we first explore injecting the point embedding at the input noise of the main and reference U-Net. In the third row, we report the results of injecting the control signal in the attention layer as introduced in \cref{sec:link}. 
In \cref{tab:enhance}, we add the enhancement strategies presented in \cref{sec:strategies} step-by-step to verify the effectiveness of our designs.

\begin{figure}[t]
\centering 
\includegraphics[width=0.9\linewidth]{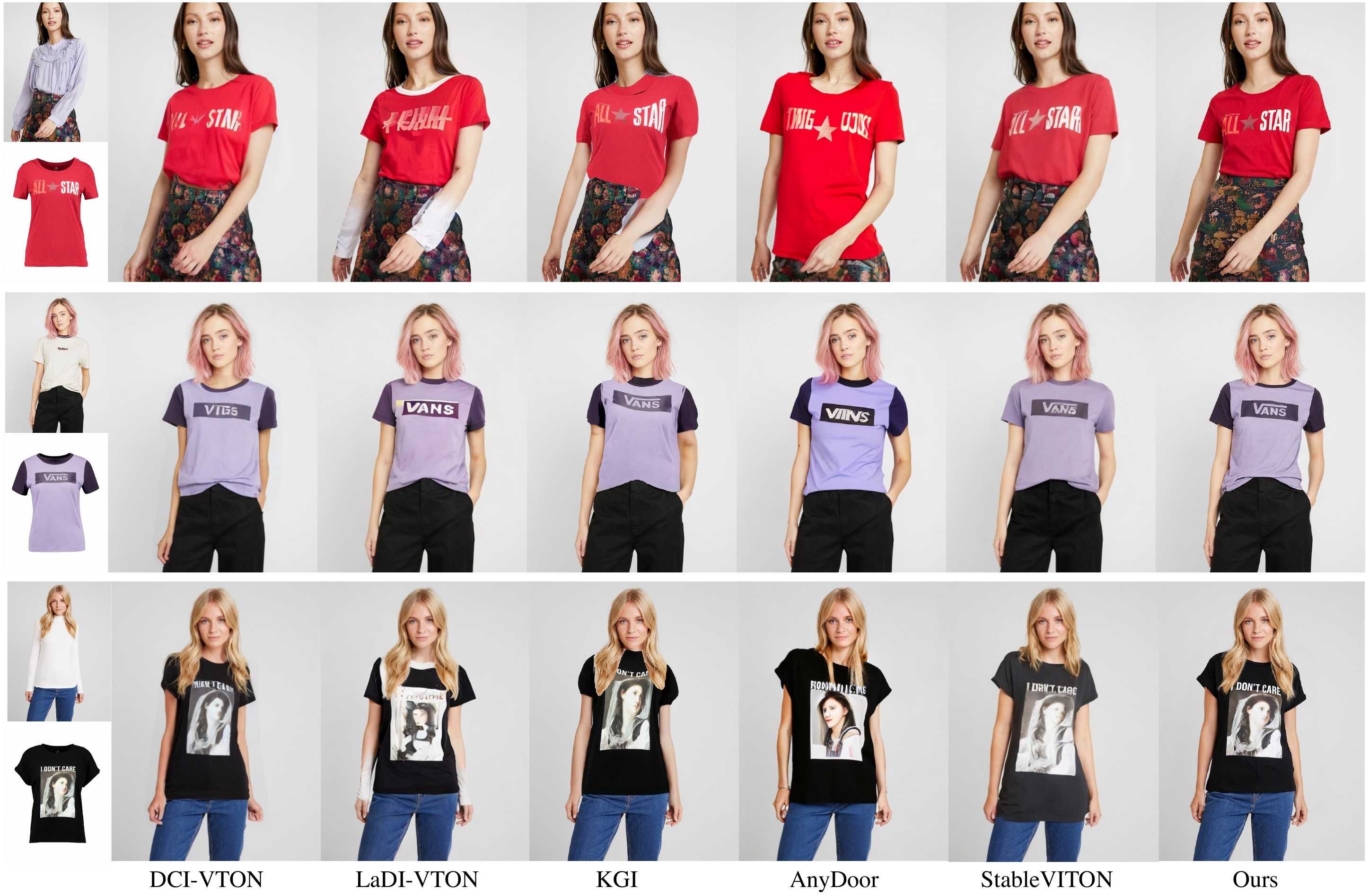} 
\vspace{-5pt}
\caption{%
    \textbf{ Qualitative comparison for classical virtual try-on.
    } 
    We make comparisons on VITON-HD~\cite{vitonhd} test split with DCI-VTON~\cite{gou2023taming}, LaDI-VTON~\cite{morelli2023ladi}, KGI~\cite{kgi}, AnyDoor~\cite{chen2023anydoor}, and StableVITON~\cite{kim2023stableviton}. Our solution demonstrates notable superiority in detail preservation and generation quality. 
}
\label{fig:tryon_compare}
\vspace{-10pt}
\end{figure}

\begin{table}[t]
\scriptsize
\begin{minipage}[c]{.49\linewidth}
    \begin{center}
    \caption{ \textbf{Point embedding injection}. We compare different ways of injecting the point embeddings and report the landmark distance as the metric.
    }   
    \vspace{-10pt}
    \label{tab:embedding_injection}
        \scalebox{0.95}{
        \begin{tabular}{lccc}
        \toprule[1pt]
                 & Dist$_\text{upper}$  & Dist$_\text{down}$ & Dist$_\text{coat}$   \\
        \hline
        None    &  35.65  & 21.13  & 43.34  \\
        Latent Noise    &  27.32  & 16.34  & 30.38  \\
        Attention q,k   &  24.35  & 15.79  & 27.27  \\
        
        \bottomrule
        \end{tabular}
        }
    \end{center}
\vspace{-0.4cm}
\end{minipage}
\hspace{2mm}
\begin{minipage}[c]{.49\linewidth}
    \begin{center}
    \caption{ \textbf{Enhancing strategies} for the sparse correspondence alignments are added step by step from the baseline to verify the effectiveness.
    }   
    \vspace{-10pt}
    \label{tab:enhance}
        \scalebox{0.8}{
        \begin{tabular}{lccc}
        \toprule[1pt]
                 & Dist$_\text{upper}$  & Dist$_\text{down}$ & Dist$_\text{coat}$   \\
        \hline
        Base~(Attention k,q)  &  24.35  & 15.79  & 27.27  \\
        + Zero-init   & 22.65   & 15.33  & 25.56  \\
        + Condition-dropping  &  18.39  & 12.04  & 20.44  \\
        + Point-weighted loss  &  17.65  & 10.32  & 20.32  \\
        \bottomrule
        \end{tabular}
        }
    \end{center}
\vspace{-0.4cm}
\end{minipage}
\vspace{-10pt}
\end{table}

\noindent \textbf{Training point pair collection.} As introduced in \cref{sec:collect}, we collect the control point by using a siamese U-Net structure. In this section, we make extensive experiments by comparing different correspondence-matching methods. 
We also utilize the benchmarks of fashion landmarks to evaluate the matching accuracy. Concretely, given a pair of landmarks on the person and garment image, we leverage different matching methods to map the landmarks from the person image to the garment image. Then, we calculate the distance between the mapped results with the ground truth landmarks.
The comparison results are listed in \cref{tab:abl_match}. The pre-trained diffusion models demonstrated superior abilities compared with other feature extractors like CLIP~\cite{CLIP} and DINOv2~\cite{dinov2}.  We also include several specific correspondence-matching methods, and our pre-trained reference U-Net.
Among them, the diffusion matcher demonstrates the best performance.

\subsection{Comparisons with Existing Alternatives}
In this section, we conduct intensive comparisons with existing alternatives. Considering that no methods could accomplish the manipulable virtual try-on. We first compete for previous arts in the standard setting of virtual try-on. Afterward, we compare interactive image editing methods that support click/drag.  

\noindent \textbf{Standard virtual try-on.} 
We report the qualitative results in \cref{tab:tryon_compare} on VITON-HD~\cite{vitonhd} and Dresscode~\cite{morelli2022dress} datasets. Our model gets the best result for the FID and KID and competitive performance for the SSIM and LPIPS. 

Considering that the quantitative results could not perfectly align with the real generation quality. 
We make qualitative comparisons with previous state-of-the-art solutions in \cref{fig:tryon_compare}. \method demonstrates obvious advantages over other works for the generation quality and detail preservation.

\begin{figure}[t]
\centering 
\includegraphics[width=0.98\linewidth]{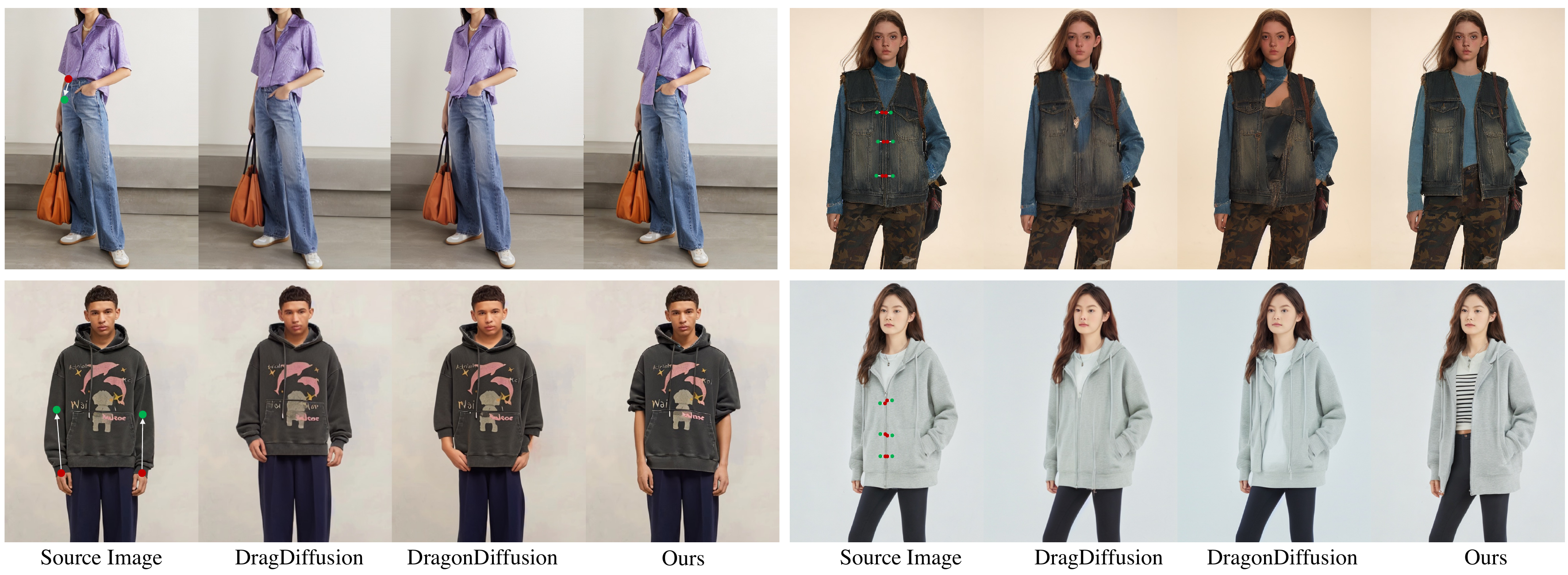} 
\vspace{-5pt}
\caption{%
    \textbf{ Qualitative comparison with drag-based image editing methods.
    } 
    We make comparisons with DragDiffusion~\cite{shi2023dragdiffusion} and DragonDiffusion~\cite{mou2023dragondiffusion}
}
\label{fig:drag_compare}
\vspace{-5pt}
\end{figure}

\noindent \textbf{Controllable generation.} We prove the controllability of our model by comparing it with drag-based image editing methods like DragDiffusion~\cite{shi2023dragdiffusion} and DragonDiffusion~\cite{mou2023dragondiffusion}. We illustrate the comparison results in \cref{fig:drag_compare}. We observe that DragDiffusion~\cite{pan2023drag} could not precisely follow the instructions of drag, while DragonDiffusion~\cite{mou2023dragondiffusion} usually destroys the structure of humans and garments.

\begin{table}[t]
\caption{
    \textbf{Ablation studies for training point-pairs collection. } 
     We compare the features of different vision foundation models and specific dense matching methods. 
    We compare the landmark distance for evaluation.
    }
\label{tab:abl_match}
\vspace{-10pt}
\centering\footnotesize
\setlength{\tabcolsep}{11pt}
\scalebox{0.75}{
    \begin{tabular}{lccc}
    \toprule[1pt]
        & Dist$_\text{upper}$  & Dist$_\text{down}$ & Dist$_\text{coat}$   \\
    \hline
    SuperGlue~\cite{sarlin2020superglue} &  134.04  & 128.34  & 187.30  \\
    CLIP~\cite{CLIP}  &  93.42  & 89.23  & 129.24  \\
    DINO~\cite{dinov2} &  83.24  & 70.08  & 103.54  \\
    \hline
    Reference U-Net   &  59.34  & 35.23  & 79.98  \\
    Stable Diffusion   &  43.44  & 29.94  & 59.45  \\
    \bottomrule
    \end{tabular}
    }
\vspace{-5pt}
\end{table}

\begin{figure}[t]
\centering 
\includegraphics[width=0.9\linewidth]{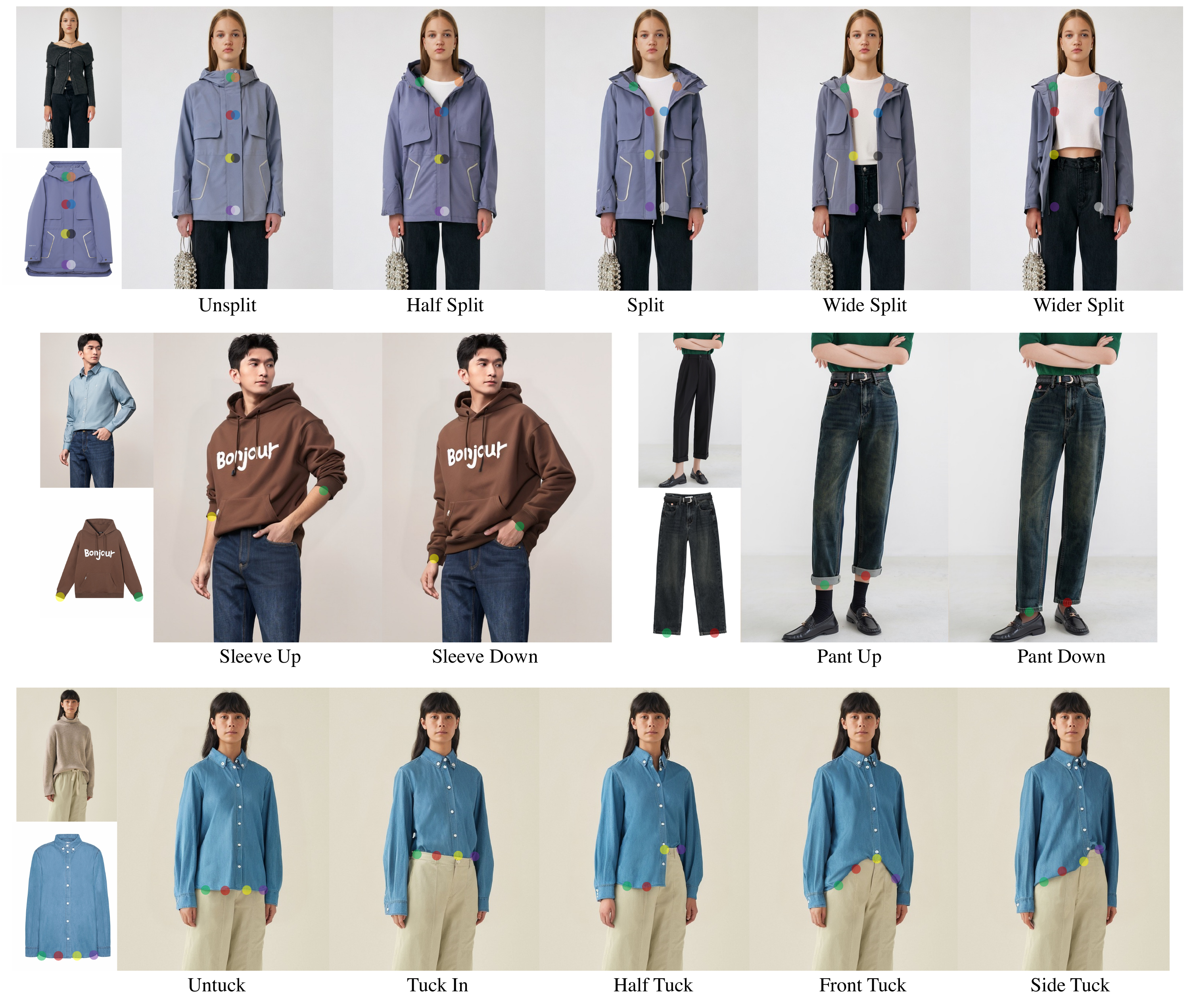} 
\vspace{-10pt}
\caption{%
    \textbf{Manipulable virtual try-on with click.} 
    \method supports users to assign arbitrary numbers of control points on the garment and person image to customize the generation, bringing diverse potentials for real-world applications.
}
\label{fig:qualitative_click}
\vspace{-5pt}
\end{figure}

\begin{table*}[!t]
\centering
\caption{Quantitative comparisons with existing state-of-the-art try-on solutions on VITON-HD and DressCode upper-body (D.C. Upper) datasets. \textbf{Bold} and \underline{underline} denote the best and the second best result, respectively.}
\vspace{-10pt}
\resizebox{0.8\linewidth}{!}{
\begin{tabular}{lcccclcccc} 
\toprule
Train / Test   &
\multicolumn{4}{c}{VITON-HD / VITON-HD} &  & \multicolumn{4}{c}{D.C. Upper / D.C. Upper}              \\ 
\cline{2-5}\cline{7-10}
 Method &
SSIM  & LPIPS & FID  & KID&  &
SSIM  & LPIPS  & FID  & KID  \\ 
\hline
VITON-HD~\cite{honda2019viton} &
0.862          & 0.117           & 12.117         & 3.23          &  &
-              & -               & -              & -              \\
HR-VITON~\cite{lee2022high} &
0.878          & 0.1045          & 11.265         & 2.73          &  &
\underline{0.936}          & 0.0652          & 13.820         & 2.71           \\
LADI-VTON~\cite{morelli2023ladi} &
0.864          & 0.0964          & 9.480          & 1.99          &  &
0.915          & 0.0634          & 14.262         & 3.33           \\
Paint-by-Example~\cite{yang2023paint} &
0.802          & 0.1428          & 11.939         & 3.85          &  &
0.897          & 0.0775          & 15.332         & 4.64           \\
DCI-VTON~\cite{gou2023taming} &
\underline{0.880 }         & \underline{0.0804}        & 8.754          & 1.10          &  &
\textbf{0.937} & \underline{0.0421}          & 11.920         & 1.89           \\
GP-VTON~\cite{xie2023gp} &
\textbf{0.884}          & 0.0814          & 9.072          & \underline{0.88  }        &  &
0.769          & 0.2679          & 20.110         & 8.17           \\ 
AnyDoor~\cite{chen2023anydoor} & 0.821   & 0.099          & 10.846         & 2.46     &   & 
0.899        & 0.119         & 14.834        & 3.05         \\ 
StableVITON~\cite{kim2023stableviton} &
0.852          & 0.0842          & \underline{8.698  }       & \underline{0.88 }         &  &
0.911          & 0.0500          & \textbf{11.266 }        & \underline{0.72  }         \\
\hline
\method& 0.877   & \textbf{0.078 }         & \textbf{8.155}          & \textbf{0.78}     &  &
0.934        & \textbf{0.0409}         & \underline{11.72}        & \textbf{0.33  }       \\ 
\bottomrule
\end{tabular}}
\vspace{-0.3cm}
\label{tab:tryon_compare}
\end{table*}

\subsection{Qualitative Analysis}
We illustrate more examples in \cref{fig:qualitative_click}, and give a qualitative analysis of the point-based manipulation.  It is demonstrated that \method supports the manipulation for various types of garments including coats, T-shirts, pants, hoodies, \textit{etc.} Besides, users could assign arbitrary numbers of control points to get the customized generation results.  As shown in the first row, assisted by the precise point control, \method can conduct ``continuous'' editing for splitting the coat gradually. The high controllability of \method enables it to realize many fantastic styles of wearing, like rolled-up sleeves or paints, and the different types of tuck.

\vspace{-10pt}
\section{Conclusion}
\vspace{-10pt}
In this work, we propose \method, a novel framework for manipulable virtual try-on. Besides reaching state-of-the-art performance on the classical setting of virtual try-on, our methods enable users to customize the style of wearing by assigning control points.  Our pipeline is based on a dual U-Net design. We introduce a sparse correspondence align module to make our model customizable. 
\method serves as a practical tool for e-commerce and provides novel inspirations for the future research of virtual try-on. 

\noindent \textbf{Limitations and potential effect.} Our methods could still generate some artifacts for fine details like human hands, especially when the hands only occupy a small region in the full image. This could be improved by using higher resolutions and larger diffusion models like SD-XL~\cite{podell2023sdxl}.
\vspace{-3pt}

\bibliographystyle{splncs04}
\bibliography{main}

\end{document}